\documentclass[letterpaper, 10 pt, conference]{ieeeconf}  

\IEEEoverridecommandlockouts                              

\usepackage{graphics} 
\usepackage{epsfig} 
\usepackage{times} 
\usepackage{amsmath} 
\usepackage{amssymb}  

\usepackage{bm}
\usepackage{color}
\usepackage{xcolor}
\usepackage{cite}
\usepackage{gensymb}
\usepackage[ruled,linesnumbered]{algorithm2e}
\usepackage{booktabs}
\usepackage{multirow}
\usepackage{threeparttable}

\title{\LARGE \bf
\textbf{Multi-Session, Localization-oriented and Lightweight LiDAR Mapping Using Semantic Lines and Planes}
}

\author{Zehuan Yu, Zhijian Qiao, Liuyang Qiu, Huan Yin and Shaojie Shen
\thanks{This work was supported in part by the HKUST Postgraduate Studentship, in part by the HKUST-DJI Joint Innovation Laboratory, and in part by the Hong Kong Center for Construction Robotics (InnoHK center supported by Hong Kong ITC).}
\thanks{Zehuan Yu, Zhijian Qiao, Huan Yin and Shaojie Shen are with the Department of Electronic and Computer Engineering, The Hong Kong University of Science and Technology, Hong Kong, China. E-mail: zyuay@connect.ust.hk, zqiaoac@connect.ust.hk, eehyin@ust.hk, eeshaojie@ust.hk. Liuyang Qiu is with the Harbin Institute of Technology (Shenzhen). E-mail: qiuliuyang11@gmail.com}
\thanks{Corresponding author: Huan Yin}
}

\begin{document}

\maketitle
\thispagestyle{empty}
\pagestyle{empty}

\begin{abstract}
In this paper, we present a centralized framework for multi-session LiDAR mapping in urban environments, by utilizing lightweight line and plane map representations instead of widely used point clouds. The proposed framework achieves consistent mapping in a coarse-to-fine manner. Global place recognition is achieved by associating lines and planes on the Grassmannian manifold, followed by an outlier rejection-aided pose graph optimization for map merging. Then a novel bundle adjustment is also designed to improve the local consistency of lines and planes. In the experimental section, both public and self-collected datasets are used to demonstrate efficiency and effectiveness. Extensive results validate that our LiDAR mapping framework could merge multi-session maps globally, optimize maps incrementally, and is applicable for lightweight robot localization. 
\end{abstract}

\section{Introduction}

Light Detection and Ranging (LiDAR) sensors are becoming standard equipment for mobile robots in the last two decades. Today's LiDAR mapping approaches, such as simultaneous localization and Mapping (SLAM), can provide dense 3D point cloud maps~\cite{zhang2014loam,chen2022direct}. Such maps have enabled various robotic navigation applications, particularly in GPS-unfriendly environments. With the deployment of multiple mobile robots for long-term autonomy, multiple LiDAR maps are constructed from multi-session data and it poses a challenge to researchers: a geometrically consistent global map is desired for robotic teams to operate effectively.

To address this challenge, researchers have designed advanced LiDAR mapping systems in a distributed~\cite{lajoie2020door,huang2021disco} or centralized~\cite{chang2022lamp,cramariuc2022maplab} manner. These systems could build a global point cloud map by integrating a chain of methods and techniques, such as place recognition, local matching, and graph optimization. These existing works lead multi-robot multi-session LiDAR mapping to a level of success, e.g., underground exploration in complex environments~\cite{ebadi2022present}. However, almost all multi-session or multi-robot mapping systems use dense feature points or point clouds. These maps bring underlying problems for robot localization and mapping, such as processing such maps on a cheap server and communication on resource-constrained vehicles, especially in a large-scale environment. One potential solution is to compress or downsample a point cloud map~\cite{labussiere2020geometry,yin20203d}. But it is hard to guarantee map consistency when a new session is incrementally merged into the reduced point cloud map.

\begin{figure}[t]
	\centering
	\includegraphics[width=\linewidth]{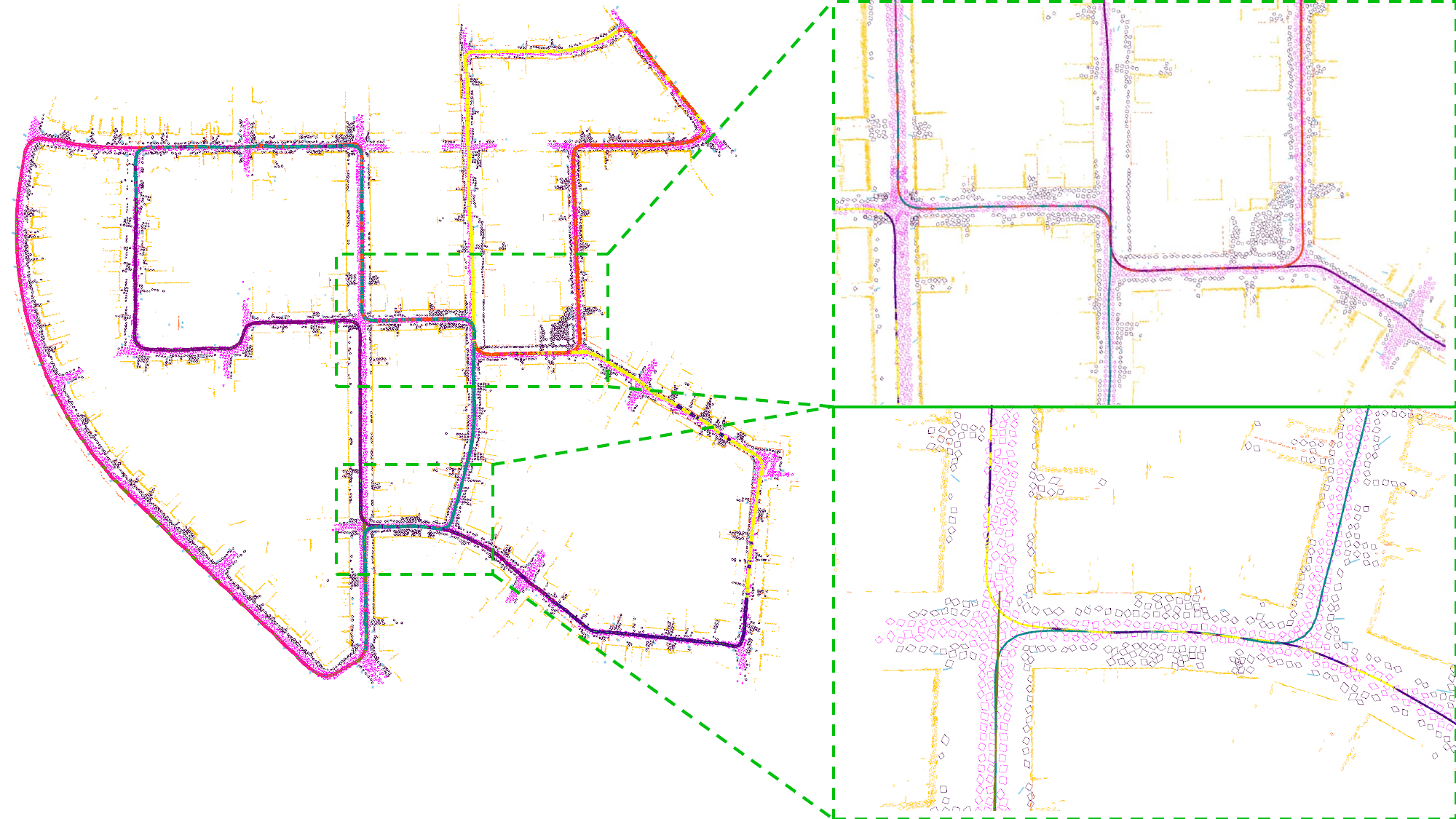}
	\caption{Mapping results generated from our proposed framework. Multi-session trajectories are with different colors in the zoom-in view. Only semantic-aided lines and planes are utilized to achieve this global and local consistent map.}
	\label{figure:cover}
\end{figure}

In this study, we propose a centralized LiDAR mapping framework for multi-session LiDAR mapping by using lightweight lines and planes rather than conventional point clouds. We carefully design a centralized LiDAR mapping framework based on lines and planes. Specifically, we build a map merging method as place recognition on the Grassmannian manifold~\cite{lusk2022global}, which unifies the data associations of lines and planes. We then conduct map optimization in a coarse-to-fine manner: first pose graph optimization followed by global bundle adjustment. Note that both global map merging and refinement use lightweight lines and planes without dense representations. Finally, the generated consistent map is lightweight  while maintaining the accuracy of map-based localization.

Overall, the primary contributions of this study can be summarized as follows:
\begin{itemize}
    \item We present a lightweight, consistent, and multi-session LiDAR mapping framework in urban environments with lines and planes.  
    \item We design a global map merging method on the Grassmannian manifold of lines and planes, ensuring  the global consistency of pose graph optimization.
    \item We propose a novel bundle adjustment with parameterized lines and planes, improving the consistency of LiDAR mapping.
    \item The proposed framework is validated with both public dataset and self-driving simulator, as well as multi-session data in large-scale urban environments.
\end{itemize}

\section{Related Work}

\subsection{Advances in LiDAR Mapping}

One well-known LiDAR-based robot mapping technique is based on scan matching methods, like iterative closest point (ICP)~\cite{pomerleau2015review}. Scan matching generally requires dense point clouds to guarantee matching accuracy. In 2014, Zhang and Singh~\cite{zhang2014loam} proposed a feature points-based pioneering work, LOAM, which extracts local LiDAR features for odometry estimation and mapping. LOAM has inspired various LiDAR SLAM systems~\cite{shan2018lego}. LOAM and its improved versions heavily rely on front-end feature extraction modules, making it less robust when applied to challenging scenarios. One promising direction is improving feature point design for a more robust and accurate LiDAR mapping system. Pan \textit{et al.}~\cite{pan2021mulls} proposed to use principal components analysis to extract feature points, and build multi-metric least square optimization for pose estimation. Liu and Zhang~\cite{liu2021balm} developed theoretical derivatives of LiDAR bundle adjustment with line and plane features, achieving map refinement as a back-end for conventional LiDAR SLAM systems. PLC SLAM by Zhou \textit{et al.}~\cite{zhou2022mathcal} introduced planes, lines, and cylinders in the LiDAR SLAM system. Despite the geometric features mentioned above, advanced learning techniques~\cite{milioto2019rangenet++} can extract semantic information in LiDAR scans and semantics can help improve LiDAR mapping. A semantic-aided LiDAR SLAM system was presented in~\cite{li2021sa}, and semantic information was utilized for local odometry estimation and global loop closure detection. 

In this study, we propose a LiDAR mapping framework that is partially inspired by previous works, including bundle adjustment and semantic-aided mapping. However, instead of using dense feature points, we directly use sparser lines and planes for both global map merging and refinement, making the entire mapping system more lightweight.


\subsection{Multi-robot and Multi-Session Mapping}

Multi-robot and multi-session localization and mapping attracts lots of research interest in recent years~\cite{saeedi2016multiple}. There are primarily two lines of research in this area: online and distributed multi-robot mapping~\cite{huang2021disco,tian2022kimera}, and centralized map servers for handling multi-session maps such as Maplab~\cite{schneider2018maplab,cramariuc2022maplab}. DiSCo-SLAM \cite{huang2021disco} used Scan Context descriptor~\cite{kim2018scan} and ICP for loop closure detection and global scan matching. Pairwise consistent measurement set maximization (PCM)~\cite{mangelson2018pairwise} was also introduced for outlier rejection in pose graph optimization (PGO). Kimera-Multi~\cite{tian2022kimera} proposed to use metric-semantic representations to build distributed SLAM system and graduated non-convexity (GNC) was introduced for robust PGO. In centralized multi-session mapping, Maplab system in 2018~\cite{schneider2018maplab} is an open source framework for visual-inertial mapping and localization, which includes map merging, batch optimization, and loop closing.. Recent Maplab 2.0~\cite{cramariuc2022maplab} integrates multiple sensor modalities into the map server, making centralized mapping more flexible. In this study, we propose a mapping framework that focuses on designing LiDAR-based global map merging and refinement for multi-session mapping in urban environments. The goal of this work is to make LiDAR mapping more lightweight and consistent with multi-session data.

We would also like to mention the existence of semantic-based map-merging techniques. Yue \textit{et al.}~\cite{yue2020collaborative} formulated a probabilistic fusion method to merge semantic point cloud maps. Guo \textit{et al.}~\cite{guo2021semantic} proposed to use semantic descriptors and design a graph-matching method to align maps. In this study, we propose a novel approach that employs semantic information in the form of lines and planes for map merging. The proposed technique operates on the Grassmannian manifold, which is distinct from conventional Euclidean space-based methods.

\begin{figure*}[t]
	\centering
	\includegraphics[width=16cm]{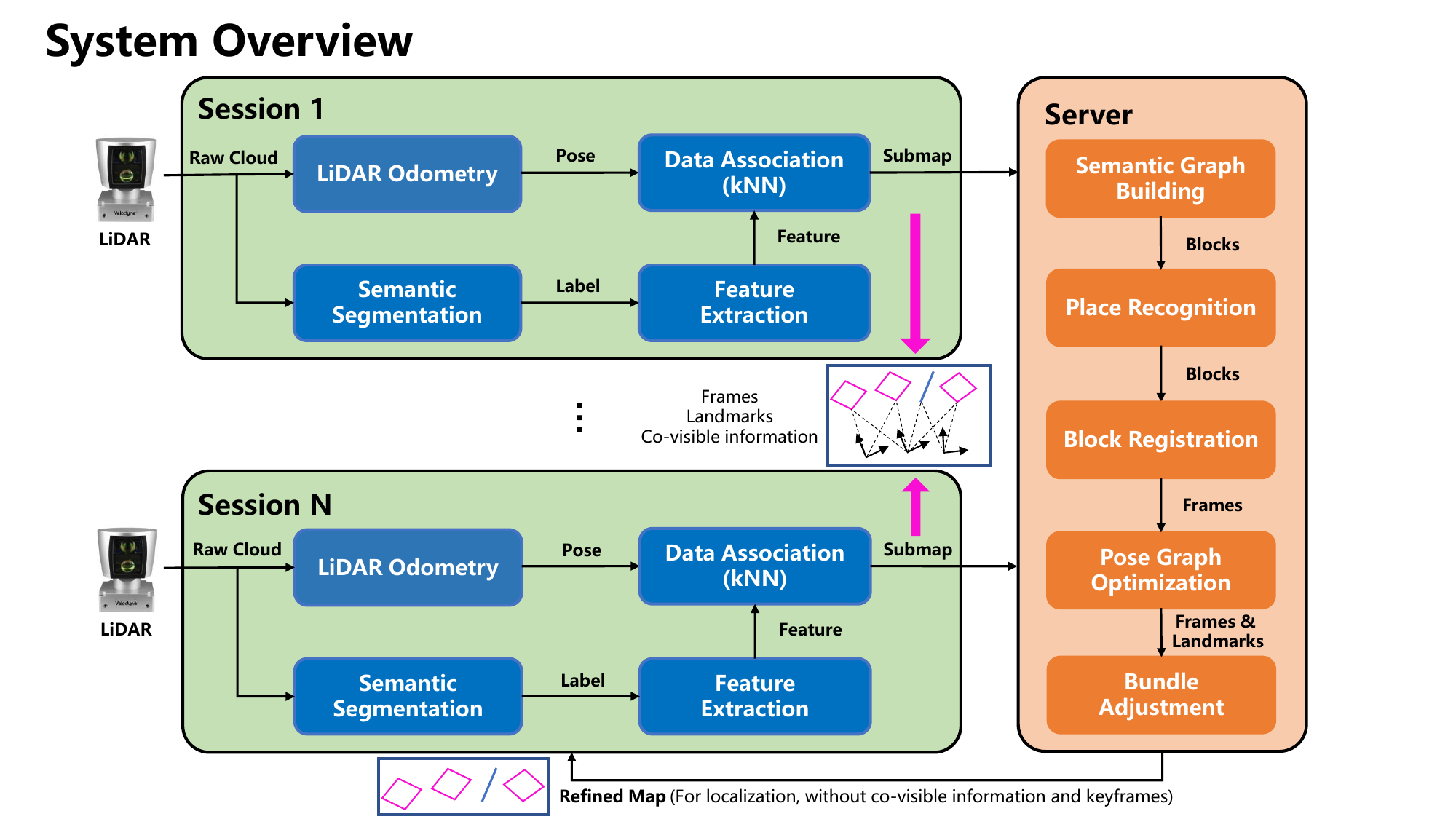}
	\caption{System Overview. Online mapping and a centralized map server are depicted in green and orange blocks, respectively. The sub-maps comprise lightweight landmarks, including lines and planes, as well as co-visibility connections between keyframes and landmarks. The map server achieves multi-session mapping from scratch in a coarse-to-fine manner, starting with global map merging and subsequently local refinement. }
	\label{figure:SystemOverview}
\end{figure*}

\section{METHODOLOGY}


An overview of the system is depicted in Figure~\ref{figure:SystemOverview}. Subsequently, this paper presents several subsections outlining the semantic mapping module using lines and planes, the proposed global map merging module, and a detailed description of the novel bundle adjustment.

\subsection{From Point Clouds to Lines and Planes}
\label{section:semantic}

\subsubsection{LiDAR Odometry and Semantic Segmentation}
The LiDAR odometry technique is well-studied in the community. In this study, we directly use it as the front end to obtain the odometry data and raw point cloud sub-maps for each session. Specifically, keyframes of sub-maps are generated based on variations in rotation and translation. We use open-source methods to obtain semantic labels of LiDAR scans. We will introduce the settings in the experimental section. The dense semantic point clouds are used for the following feature extraction and parameterization.

\subsubsection{Semantic Feature Extraction}

Several specific types of semantic landmarks, such as poles, roads, buildings, and fences, are selected as map elements based on prior knowledge of urban scenarios. These elements commonly exist in urban environments and have compact geometric representations. In this study, poles can be represented as infinite lines, while other landmarks are represented as infinite planes. To extract these line and plane features from dense semantic point clouds, a clustering algorithm \cite{ester1996density} and a voxel-based segmentation algorithm are used. Essentially, line and plane features are low-dimensional landmarks in this study. These line and plane features have fewer parameters compared to dense point clouds, while still preserving the original geometric information, making them beneficial for map management in large-scale urban environments.

Specifically, for line feature extraction, clustering results provide several point cloud clusters that described different pole-like objects. The direction and centroid of each object are obtained using the principal component analysis (PCA) algorithm. For plane features, the semantic point cloud is divided into voxels based on position and encoded using a hash function, which is from \cite{liu2021balm}. Points within each voxel are checked using the PCA algorithm, and plane features are extracted. The centroid and covariance matrix of each feature are saved in their host frame.

To build a lightweight mapping framework, a line observation is described using two terminal points $\mathbf{f}^{\mathcal{L}}:= \langle \mathbf{p}_a, \mathbf{p}_b \rangle$, which are determined using the singular values and singular vectors from the PCA result. Similarly, a plane observation is described by the centroid and four terminal points $\mathbf{f}^{\mathcal{S}}:= \langle \mathbf{p}_a, \mathbf{p}_b, \mathbf{p}_c, \mathbf{p}_d\rangle$, which also depended on the PCA result. The four points are the vertices of a rhombus and lie in the two directions of the singular vectors corresponding to the larger two singular values.

\subsubsection{Lightweight Map Structure}

We then initialize and update line and plane landmarks during the online mapping procedure, where data association based on the centroid-based nearest neighbor searching method constructs the co-visibility structure. We define a line landmark $\mathbf{l}^{\mathcal{L}}:=\langle l_s, \mathbf{c}^{\mathcal{L}}, \mathbf{n}^{\mathcal{L}},
\mathbf{p}^{\mathcal{L}},
\left\{ \mathbf{f}^{\mathcal{L}}_i \right\} \rangle$ and a plane landmark $\mathbf{l}^{\mathcal{S}}:=\langle l_s, \mathbf{c}^{\mathcal{S}}, \mathbf{n}^{\mathcal{S}},
\mathbf{p}^{\mathcal{S}},
\left\{ \mathbf{f}^{\mathcal{S}}_i \right\} \rangle$, including a semantic label, centroid, normal, minimum parameter block, and their observation in different keyframes, similar to the visual bundle structure.

However, the point-normal form, which represents normal vectors $\bf{n}$ and random points $\bf{c}$ in Euclidean space, is over-parameterized for lines and planes and is not friendly for constructing optimization problems. To address this issue, we propose to represent an infinite line as $\mathbf{p}^{\mathcal{L}}:= \langle \alpha, \beta, {x}, {y}\rangle \in\mathbb{R}^4$, where $\alpha$ and $\beta$ represent the direction, and ${x}$ and ${y}$ represent the offset translation on the $xOy$ plane. Similarly, an infinite plane is formulated as $\mathbf{p}^{\mathcal{S}}:=\langle \alpha, \beta, {d}\rangle\in\mathbb{R}^3$, where $\alpha$ and $\beta$ represent the direction, and ${d}$ represents the offset translation on the $z$-axis, which are the minimum parameter blocks mentioned before.

\begin{equation}
    \small
    \mathbf{R}\left( \alpha ,\beta \right) =\left[ \begin{matrix}
	\cos \left( \beta \right)&		0&		-\sin \left( \beta \right)\\
	\sin \left( \alpha \right) \sin \left( \beta \right)&		\cos \left( \alpha \right)&		\sin \left( \alpha \right) \cos \left( \beta \right)\\
	\cos \left( \alpha \right) \sin \left( \beta \right)&		-\sin \left( \alpha \right)&		\cos \left( \alpha \right) \cos \left( \beta \right)\\
    \end{matrix} \right] 
    \label{eq:Rotation2dof}    
\end{equation}


\begin{figure}[t]
	\centering
	\includegraphics[width=8cm]{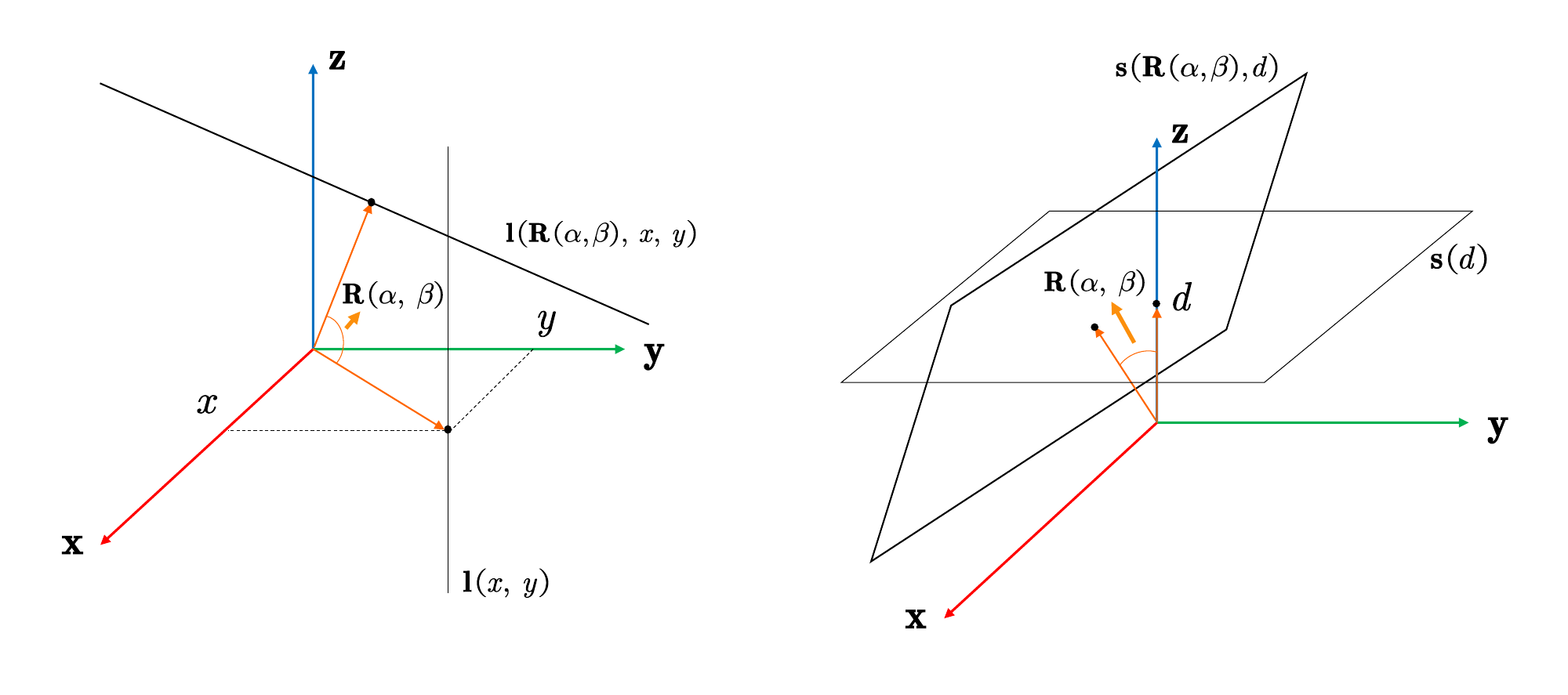}
	\caption{Formulation of line and plane landmarks.}
	\label{figure:LandmarkFormulation}
\end{figure}

In fact, it is possible to transform all infinite lines and infinite planes from the original line $\mathbf{u}_z$ and original plane $xOy$ in two steps, as demonstrated in Figure \ref{figure:LandmarkFormulation}. Regarding line features, we first apply an offset $(x, y)$ to the original line $\mathbf{u}_z$ to obtain the translated line $\mathbf{l}(x,y)$. Next, we rotate the nearest point $(x,y)$ using a 2 degrees of freedom (DoF) rotation matrix $\mathbf{R}(\alpha, \beta)$ to obtain the final line $\mathbf{l}(\mathbf{R}(\alpha, \beta), x,y)$. The 2 DoF rotation matrix $\mathbf{R}(\alpha, \beta)$ is defined by equation \eqref{eq:Rotation2dof}. Concerning plane features, the original plane $xOy$ is translated along the $z$ axis and rotated using a 2 DoF rotation matrix, resulting in the final plane $\mathbf{s}(\mathbf{R}(\alpha, \beta),z)$. 

The minimum line or plane representation is devised for the optimization framework, necessitating a residual related to the pose and landmark. While point-to-line and point-to-plane residuals are available, they are constructed using the point-normal form. Hence, it is essential to establish the mapping relations between the point-normal form and infinite line/plane form, and our proposed ones are stated as follows:
\begin{equation} \small
    \begin{bmatrix}
    \mathbf{n}^{\mathcal{L}} \\ \mathbf{c}^{\mathcal{L}}
    \end{bmatrix}
    =
    \begin{bmatrix}
        \mathbf{R}(\alpha, \beta)\mathbf{u}_z \\
        \mathbf{R}(\alpha, \beta)(\mathbf{u}_x x + \mathbf{u}_y y)
    \end{bmatrix}
    \label{eq:LineConvert}
\end{equation}
\begin{equation} \small
    \begin{bmatrix}
    \mathbf{n}^{\mathcal{S}} \\ d^{\mathcal{S}}
    \end{bmatrix}
    =
    \begin{bmatrix}
        \mathbf{R}(\alpha, \beta)\mathbf{u}_z \\
        d
    \end{bmatrix}
    \label{eq:SurfConvert}
\end{equation}

\subsection{Global Map Merging}
\label{section:GMM}

\subsubsection{Semantic Graph Building}

To merge sub-maps at different locations, place recognition and relative pose estimation are critical challenges that must be solved globally, i.e., without an initial guess. Traditional methods typically employ full laser scan data to construct handcrafted or learning-based global descriptors. In our case, we implement the GraffMatch algorithm \cite{lusk2022graffmatch}, which is a global descriptor-free method based on the open-sourced data association framework \cite{lusk2021CLIPPER} to identify the overlapping parts between two sub-maps. However, since each sub-map contains numerous landmarks, the graph matching problem has an infeasible dimensionality, leading to unmanageable solution times.

To overcome this issue, we propose a carefully designed two-step method. Firstly, we partition the sub-map into blocks along the vehicle trajectory instead of directly matching the two sub-maps. Each block comprises a host keyframe and a set of landmarks around it. Secondly, to reduce the number of features, we cluster the plane landmarks that lie on the same infinite plane into a single plane. Specifically, we represent each pole as a line node $\mathbf{v}^L=\langle \mathrm{l}_{\mathrm{s}}, \mathbf{c}, \mathbf{n}, \mathbf{l}^L \rangle$, and cluster the plane landmarks to form larger planes. These larger planes are represented as plane nodes $\mathbf{v}^S=\langle \mathrm{l}_{\mathrm{s}}, \mathbf{c}, \mathbf{n}, \left\{\mathbf{l}_i^S\right\} \rangle$. The elements of the graph node include semantic labels, centroids, normals, and the corresponding landmark(s). The semantic graph consists of two types of nodes, namely $\mathcal{G}=\langle \left\{\mathbf{v}_i^L \right\}, \left\{\mathbf{v}_j^S \right\} \rangle$.

\subsubsection{Place Recognition}

In the GraffMatch algorithm, line and plane features are extracted to formulate the $k$-dim subspaces, and their Graff coordinates ${Y}$ are shown as \eqref{eq:GraffCoordinate}.

\begin{equation} \small
    \begin{aligned}
        {Y} = \begin{bmatrix}
            A & b/\sqrt{\lVert b \rVert^2+1} \\
            0 & 1/\sqrt{\lVert b \rVert^2+1}
        \end{bmatrix} 
         \in \mathbb{R}^{(n+1)\times(k+1)}
    \end{aligned}
    \label{eq:GraffCoordinate}
\end{equation}

In our situation, $A$ and $b$ are the orthonormal basis and orthonormal displacement of the line or plane extracted from pole instances and clustered plane instances. Grassmannian metric $\mathrm{d}_{\mathrm{Graff}}({Y}_1, {Y}_2)$ can be used to compute the distance of two subspaces, such as:

\begin{equation} \small
    \begin{aligned}
    b_{02}&=b_2-b_1 \\
    Y^{'}_1 &= \begin{bmatrix}
            A_1 & 0 \\
            0 & 1
        \end{bmatrix}   \\  
    Y^{'}_2 &= \begin{bmatrix}
            A_2 & b_{02}/\sqrt{\lVert b_{02} \rVert^2 + 1} \\
            0 & 1/\sqrt{\lVert b_{02} \rVert^2 + 1}
        \end{bmatrix} \\
    \left\{\sigma_i\right\} &= \mathrm{SVD}((Y^{'}_1)^TY^{'}_2)    \\
    \mathrm{d}_{\mathrm{Graff}}({Y}_1, {Y}_2) &= \
    \sum_{i}{\mathrm{arccos}^2(\sigma_i)} 
    \end{aligned}
\end{equation}

Then this metric is used to construct the affinity matrix for the following global matching framework. We recommend interested readers refer to CLIPPER \cite{lusk2022graffmatch} for a more detailed explanation.

\begin{figure*}[t]
	\centering
	\includegraphics[width=15cm]{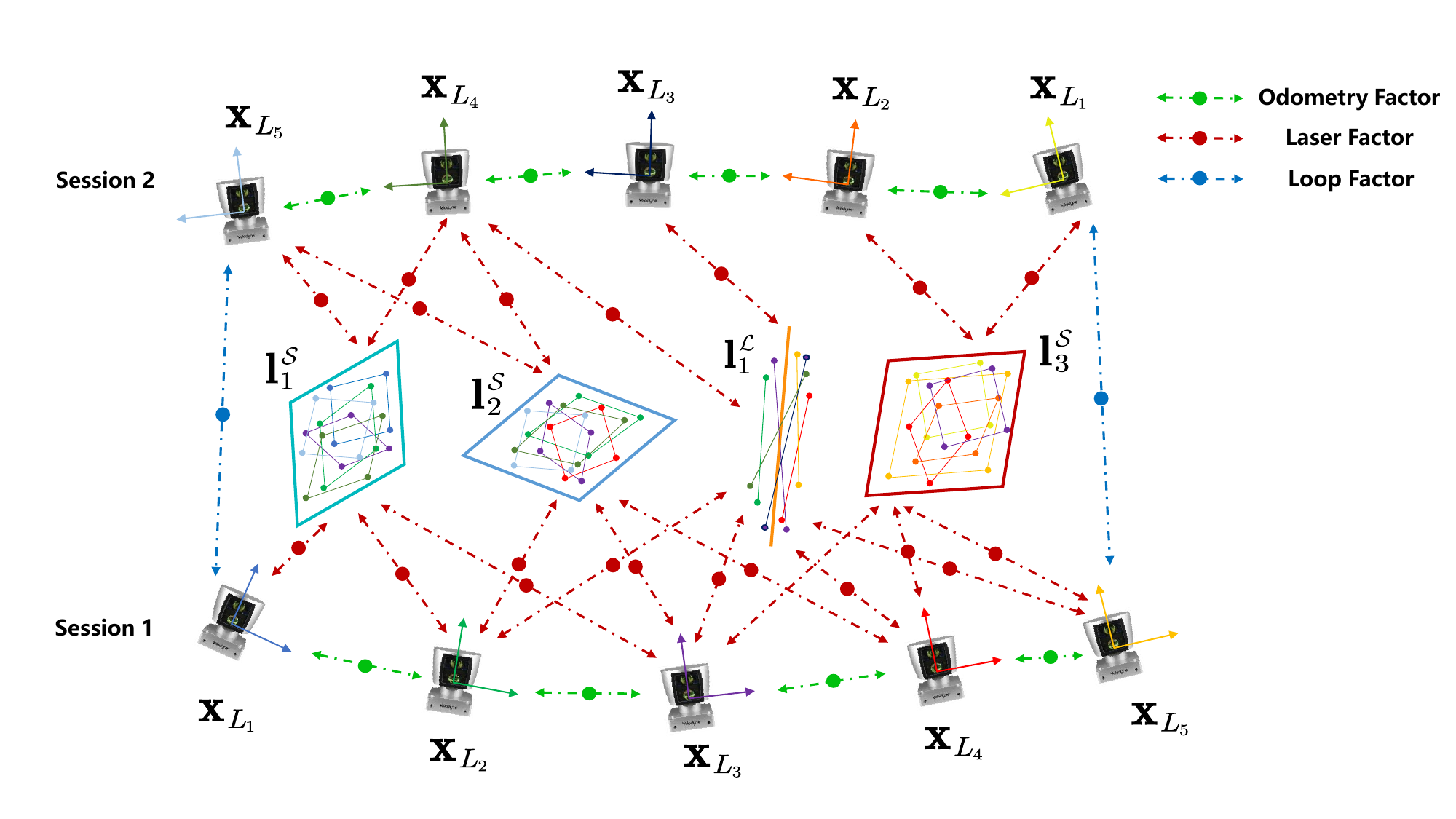}
	\caption{Landmark definition and factors for multi-session LiDAR mapping: odometry factors (in green) obtained using LOAM, loop factors (in blue) obtained using GraffMatch, and laser factors (in red) obtained through the proposed bundle adjustment. Distinct colors are employed for lines and planes to enhance the description, with respect to its connected robot pose in the two sessions.}
	\label{figure:LaserFactorGraph}
\end{figure*}

\subsubsection{Block Registration}

Thanks to the CLIPPER framework, which gives the corrected line and plane correspondences $\mathcal{C}^{\mathcal{L}}$ and $\mathcal{C}^{\mathcal{S}}$ between semantic graphs $\mathcal{G}_{b_i}$ and $\mathcal{G}_{b_j}$ (${b_i}$ and ${b_j}$ are two individual blocks). Then we can build a hybrid registration with lines and planes to solve the relative pose, which is formulated as 

\begin{equation} \small
    \begin{aligned}
        \mathop{\mathrm{min}}\limits_{\mathbf{T}^{b_i}_{b_j}}
        & \sum_{(k, l)\in \mathcal{C}^{\mathcal{L}}} \rho( \lVert (\mathbf{I} - \mathbf{n}_k\mathbf{n}^T_k)(\mathbf{T}^{b_i}_{b_j} \mathbf{c}_l - \mathbf{c}_k) \rVert_{\Sigma_{\mathcal{L}}}^{2}) \\
        & \sum_{(k, l)\in \mathcal{C}^{\mathcal{S}}} \rho( \lVert \mathbf{n}^T_k(\mathbf{T}^{b_i}_{b_j} \mathbf{c}_l - \mathbf{c}_k) \rVert_{\Sigma_{\mathcal{S}}}^{2})
    \end{aligned}
    \label{eq:BlockCoarseRegistraion}
\end{equation}

where $\mathbf{T}^{b_i}_{b_j}$ is the relative transformation between block $i$ and block $j$, while $\rho(\cdot)$ denotes a robust kernel function. However, due to the limitations of CLIPPER, it may not always be possible to find all the corresponding node pairs. Therefore, to obtain a more accurate estimate of the relative pose, an optimization-based refinement based on knn-search-based iterative closest landmark is performed, which is formulated as follows:

\begin{equation} \small
    \begin{aligned}
        \mathop{\mathrm{min}}\limits_{\mathbf{T}^{f_i}_{f_j}}
        & \sum_{(k, l)\in \mathcal{N}^{\mathcal{L}}} \rho( \lVert (\mathbf{I} - \mathbf{n}_k\mathbf{n}^T_k)(\mathbf{T}^{f_i}_{f_j} \mathbf{p}_l - \mathbf{p}_k) \rVert_{\Sigma_{\mathcal{L}}}^{2}) \\
        & \sum_{(k, l)\in \mathcal{N}^{\mathcal{S}}} \rho( \lVert \mathbf{n}^T_k(\mathbf{T}^{f_i}_{f_j} \mathbf{p}_l - \mathbf{p}_k) \rVert_{\Sigma_{\mathcal{S}}}^{2})
    \end{aligned}
    \label{eq:BlockRefineRegistraion}
\end{equation}

where $(k,l)$ is the nearest-neighbor pair between two blocks. To achieve a more robust registration result, the optimization problem is solved iteratively for improved matching of landmarks, leading to a more stable solution. Compared to descriptor-based place recognition methods such as Scan Context \cite{kim2018scan}, GraffMatch is descriptor-free but computationally more intensive. Once all pairs of blocks have been registered, we use the pairwise consistent measurement (PCM) algorithm \cite{mangelson2018pairwise} to identify false loop candidates. The function $C(\mathbf{T}^{j_k}_{i_k}, \mathbf{T}^{j_l}_{i_l})$ measures the consistency of the relative pose of the block registration:


\begin{equation} \small
    \begin{aligned}
    \delta \mathbf{T} &= (\mathbf{T}^{j_k}_{i_k})^{-1} \cdot \mathbf{\hat T}^{i_l}_{i_k} \cdot \mathbf{T}^{j_l}_{i_l} \cdot \mathbf{\hat T}^{j_k}_{j_l} \\
        C(\mathbf{T}^{j_k}_{i_k}, \mathbf{T}^{j_l}_{i_l}) &= 
        [\mathbf{r}(\delta\mathbf{R}), \mathbf{r}(\delta \mathbf{p})]^T\\
        &= [ \lVert \mathrm{Log}(\mathbf{R}(\delta\mathbf{T})) \rVert_2,  
        \lVert \mathbf{p}(\delta\mathbf{T}) \rVert_2]^T
    \end{aligned}
\end{equation}

If $\mathbf{r}(\delta\mathbf{R})$ and $\mathbf{r}(\delta \mathbf{p})$ are small enough, the pair of relative pose $\mathbf{T}^{j_k}_{i_k}$ and $\mathbf{T}^{j_k}_{i_k}$ are considered to be consistent. The problem of solving the internal-consistent set of relative poses $\mathcal{L}_\mathcal{O}$ is a maximum clique problem that can be solved by the open-source parallel maximum clique library~\cite{rossi2013fast}, such that all the remaining relative poses will be constructed as relative pose residual~\eqref{eq:RelPoseResidual} and added to the following pose graph optimization. 

\begin{equation} \small
\begin{aligned}
    \mathbf{r}_{\mathcal{L}_\mathcal{O}}\left(\mathbf{T}^W_{L_k}, \mathbf{T}^W_{L_{k+1}}, \mathbf{\hat T}^{L_{k}}_{L_{k+1}}\right)=\\
    \begin{bmatrix}
    {\mathbf{R}^W_{L_k}}^T \left( \mathbf{p}^W_{L_{k+1}} - \mathbf{p}^W_{L_{k}} \right) - \mathbf{\hat p}^{L_{k}}_{L_{k+1}} \\
    (\mathbf{\hat R}^{L_k}_{L_{k+1}})^T (\mathbf{R}^W_{L_k})^T \mathbf{R}^W_{L_{k+1}}
    \end{bmatrix}
    \label{eq:RelPoseResidual}    
\end{aligned}
\end{equation}

\subsection{Map Refinement}

\subsubsection{Pose Graph Optimization}

Generally, based on the front-end LiDAR odometry data, the odometry pose residuals $\mathbf{r}_{\mathcal{O}}$ have the same formulation as $\mathbf{r}_{\mathcal{L}_\mathcal{O}}$. Thus, the classical pose graph optimization problem is formulated as

\begin{equation} \small
\begin{aligned}
    \mathop{\mathrm{min}}\limits_{\mathbf{T}^W_{L_i} \in \mathcal{F}} 
    & \sum_{(k, k+1)\in \mathcal{O}} \rho( \lVert \mathbf{r}_{\mathcal{O}}\left(\mathbf{T}^W_{L_k}, \mathbf{T}^W_{L_{k+1}}, \mathbf{\hat T}^{L_{k}}_{L_{k+1}}\right) \rVert_{\Sigma_{\mathcal{O}}}^{2}) + \\ & \sum_{(i, j)\in \mathcal{L}_{\mathcal{O}}} \rho( \lVert \mathbf{r}_{\mathcal{L}_\mathcal{O}}\left(\mathbf{T}^W_{L_i}, \mathbf{T}^W_{L_{j}}, \mathbf{\hat T}^{L_{i}}_{L_{j}}\right) \rVert_{\Sigma_{\mathcal{L}_\mathcal{O}}}^{2})
    \label{eq:LBA}
\end{aligned}
\end{equation}

Pose graph optimization provides a higher-precision global pose for keyframes and landmarks. However, it is possible to have landmarks that are repeatedly included in multiple sub-maps. To reduce the size of the map and the dimension of subsequent optimization, the instances of these landmarks in multiple sub-maps will be merged depending on either the graph-matching result or the centroid distance.

\subsubsection{Bundle adjustment for lines and planes}
\label{section:localBA}

After merging the overlap landmarks between sub-maps, we introduce a new bundle adjustment formulation to jointly optimize the keyframes' poses, line landmarks, and plane landmarks to improve the map accuracy.

Inspired by visual bundle adjustment methods, we construct the residual between the observation in the keyframe and the corresponding landmark in the world. Every line observation has two point-to-infinite-line residuals, which can be expressed as follows:
\begin{equation} \small
    \mathbf{r}_{\mathcal{L}}\left( \mathbf{T}^W_L, \mathbf{l}^{\mathcal{L}} \right) =\left( \mathbf{I}-\mathbf{nn}^T \right) \left( \mathbf{T}^W_L\mathbf{p}^{L}-\mathbf{q} \right) 
    \label{eq:LineResidual}
\end{equation}
where $\mathbf{T}^W_L$ is the pose of the keyframe, $\mathbf{l}^{L}$ is the line landmark whose point-normal form is $\left( \mathbf{n}, \mathbf{q} \right)$, and $\mathbf{p}_{L}$ represents one of the two observation points in the keyframe. 



In the same way, every plane observation has four point-to-infinite-plane residuals, such as
\begin{equation} \small
    \mathbf{r}_{\mathcal{S}}\left( \mathbf{T}^W_L, \mathbf{l}^{\mathcal{S}} \right) =\mathbf{n}^T\left( \mathbf{T}^W_L\mathbf{p}^{L}-\mathbf{q} \right)
    \label{eq:SurfaceResidual}
\end{equation}
With \eqref{eq:LineConvert}, \eqref{eq:SurfConvert}, \eqref{eq:LineResidual} and \eqref{eq:SurfaceResidual}, we could easily calculate the jacobian matrices with respect to line parameters $\langle\alpha, \beta, x, y\rangle$ and plane parameters $\langle \alpha, \beta, d\rangle$. Then we formulate the LiDAR bundle adjustment formulation as 
\begin{equation}
\small
\begin{aligned}
    \mathop{\mathrm{min}}\limits_{\mathcal{X}} 
    \sum_{(k, k+1)\in \mathcal{O}} &\rho( \lVert \mathbf{r}_{\mathcal{O}}\left(\mathbf{T}^W_{L_k}, \mathbf{T}^W_{L_{k+1}}, \mathbf{\hat T}^{L_{k}}_{L_{k+1}}\right) \rVert_{\Sigma_{\mathcal{O}}}^{2}) +\\ \sum_{(i, j)\in \mathcal{L}_{\mathcal{O}}} &\rho( \lVert \mathbf{r}_{\mathcal{L}_\mathcal{O}}\left(\mathbf{T}^W_{L_i}, \mathbf{T}^W_{L_{j}}, \mathbf{\hat T}^{L_{i}}_{L_{j}}\right) \rVert_{\Sigma_{\mathcal{L}_\mathcal{O}}}^{2}) + \\
    \sum_{(k, i)\in \mathcal{C}_{\mathcal{L}}} &\rho( \lVert \mathbf{r}_{\mathcal{L}}\left( \mathbf{T}^W_{L_k}, \mathbf{l}^{\mathcal{L}}_i \right) \rVert_{\Sigma_{\mathcal{L}}}^{2}) + \\ 
    \sum_{(k, j)\in \mathcal{C}_{\mathcal{S}}} &\rho( \lVert \mathbf{r}_{\mathcal{S}}\left( \mathbf{T}^W_{L_k},\mathbf{l}^{\mathcal{S}}_i \right) \rVert_{\Sigma_{\mathcal{S}}}^{2})
    \label{eq:LBA}
\end{aligned}
\end{equation}
where $\mathcal{X}$ includes all the keyframes' poses and landmarks.

Finally, all the multi-session poses and landmarks are optimized jointly in our proposed mapping framework.  It is worth noting that the pipeline of global map merging and refinement only involves lightweight lines and planes. The effectiveness and efficiency of this map type will be validated in the following experimental sections.



\section{Experiments}

We first introduce the experimental settings and then present the extensive mapping results qualitatively and quantitatively.

\subsection{Experimental settings}

We implement our framework in C++ and employ the Ceres Solver~\cite{Agarwal_Ceres_Solver_2022} to solve the nonlinear optimization problems. To assess the efficacy of our framework, we select the KITTI dataset~\cite{geiger2012we}, a real-world dataset, and the CARLA simulator~\cite{dosovitskiy2017carla}, a virtual dataset. Both datasets provide a large number of semantic-aided scans and ground truth poses that can be used to build and evaluate our mapping framework.

For the KITTI dataset, we choose Seq.00, Seq.05, and Seq.08, which belong to urban scenarios and have several overlapping parts to generate multi-session data. We use an open-source 3D semantic segmentation method\footnote{https://github.com/mit-han-lab/spvnas}~\cite{tang2020searching} to obtain semantic labels in KITTI datasets. For one sequence in KITTI, we divide it into multiple sessions and each session is with an overlapping area with some of the other sessions. As for CARLA, the multi-session data collection process in the simulator is convenient, and the number of sessions is not limited. We use Town03 in CARLA as the data collection environment, and generate ten driving sessions with random start locations. Since users can extract all precise LiDAR data from CARLA, we add Gaussian noise to the point cloud with a variance $\sigma$ = 4cm. Additionally, 20$\%$ of the semantic labels are randomly altered to incorrect ones to create noise. For both the KITTI dataset and CARLA, we employ LOAM\footnote{https://github.com/HKUST-Aerial-Robotics/A-LOAM}~\cite{zhang2014loam} a classical LiDAR mapping framework, as the front end.


\subsection{Effectiveness of Bundle Adjustment}

We first present an individual validation of the effectiveness of our bundle adjustment method in improving local mapping consistency. Since our sub-maps do not have internal loops, we evaluate the consistency of the map by comparing the relative pose error (RPE) between the raw trajectory obtained from A-LOAM and the optimized trajectory obtained from bundle adjustment that uses the initial odometry pose from A-LOAM.

We use the KITTI Seq.00 dataset for validation, which is divided into six sessions. Our bundle adjustment algorithm could help the relative pose error of each trajectory, as shown in Table \ref{tab:KittiLaserBARPE}, which means that the local map consistency is improved. Actually, local map consistency is essential for map merging tasks since an internally inconsistent map will destroy the source map structure during merging. In our experiments, we observed that A-LOAM is not always stable when the vehicle experiences large acceleration or rotation velocity, leading to large odometry drift. Our proposed bundle adjustment could effectively correct the drift and reduce inconsistency.

\begin{table*}[t]
    \centering
    \caption{ RMSE of the RPE ($\degree$/$m$) On KITTI Seq. 00 Dataset}
    \resizebox{0.80\textwidth}{!}{
    \begin{tabular}{cccccccc}
    \toprule
    Methods & Session 0 & Session 1 & Session 2 & Session 3 & Session 4 & Session 5\\
    \midrule
    A-LOAM & 0.1547/0.0223 & 0.1318/0.0212 & 0.1433/0.0232 & 0.1260/0.0229 & 0.1318/0.0220 & 0.1375/0.0235 \\
    A-LOAM+ Ours BA & \textbf{0.1489}/\textbf{0.0195} & \textbf{0.1318}/\textbf{0.0193} & \textbf{0.1317}/\textbf{0.0220} & \textbf{0.1203}/\textbf{0.0202} & \textbf{0.1260}/\textbf{0.0197} & \textbf{0.1260}/\textbf{0.0215} \\
    \bottomrule
    \end{tabular}
    }
    \label{tab:KittiLaserBARPE}
\end{table*}

\begin{table}[!t]
    \centering
    \caption{ RMSE of the APE ($\degree$/$m$) On KITTI and CARLA Datasets}
    \resizebox{0.45\textwidth}{!}{
    \begin{tabular}{ccccc}
    \toprule
    Methods & Seq. 00 & Seq. 05 & Seq. 08 & Carla T3 \\
    \midrule
    SC-A-LOAM & 1.0/1.3 & 0.7/0.8 & 2.6/4.3 & / \\
    CT-ICP\cite{dellenbach2022ct} & \textbf{0.7}/1.7 & \textbf{0.5}/0.8 & 1.2/2.5 & / \\
    MULLS\cite{pan2021mulls} & \textbf{0.7}/1.1 & 0.6/1.0 & 1.3/2.9 & / \\
    LiTAMIN2\cite{yokozuka2021litamin2} & 0.8/1.3 & 0.7/0.6 & \textbf{0.9}/\textbf{2.1} & / \\
    LiDAR-BA\cite{liu2023large} & \textbf{0.7}/0.8 & \textbf{0.5}/\textbf{0.4} & 1.3/2.7 & / \\
    \midrule
    Ours(w/o BA) & \textbf{0.7}/0.9 & 0.6/0.6 & 2.7/4.5 & 0.229/0.512 \\
    Ours(w BA) & \textbf{0.7}/\textbf{0.7} & 0.6/0.5 & 2.7/4.5 & \textbf{0.143}/\textbf{0.432}\\
    \bottomrule
    \end{tabular}
    }
    \label{tab:MultiSessionAPE}
\end{table}

\begin{table}[!t]
    \centering
    \caption{Map Storage (MB) of Different Representations }
    \resizebox{0.45\textwidth}{!}{
    \begin{tabular}{cccc}
    \toprule
    Representation & KITTI Seq. 00 & KITTI Seq. 05 & KITTI Seq. 08 \\
    \midrule
    Cloud ($r$=0.1m) & 1773.40 & 1236.75 & 1254.92 \\
    Cloud ($r$=0.3m) & 23.48 & 16.77 & 18.56 \\
    Cloud ($r$=0.5m) & 8.33 & 6.01 &  6.57 \\
    \midrule
    Ours & 1.85 (0.50/km) & 1.22 (0.55/km) & 1.56 (0.48/km) \\
    Ours (L) & \textbf{0.77} (0.21/km) & \textbf{0.52} (0.24/km) & \textbf{0.64} (0.20/km) \\
    \bottomrule
    \end{tabular}
    }
    \label{tab:MapStorage}
\end{table}

\subsection{Multi-Session map merging and consistency}

\begin{figure}[t]
	\centering
	\includegraphics[width=\linewidth]{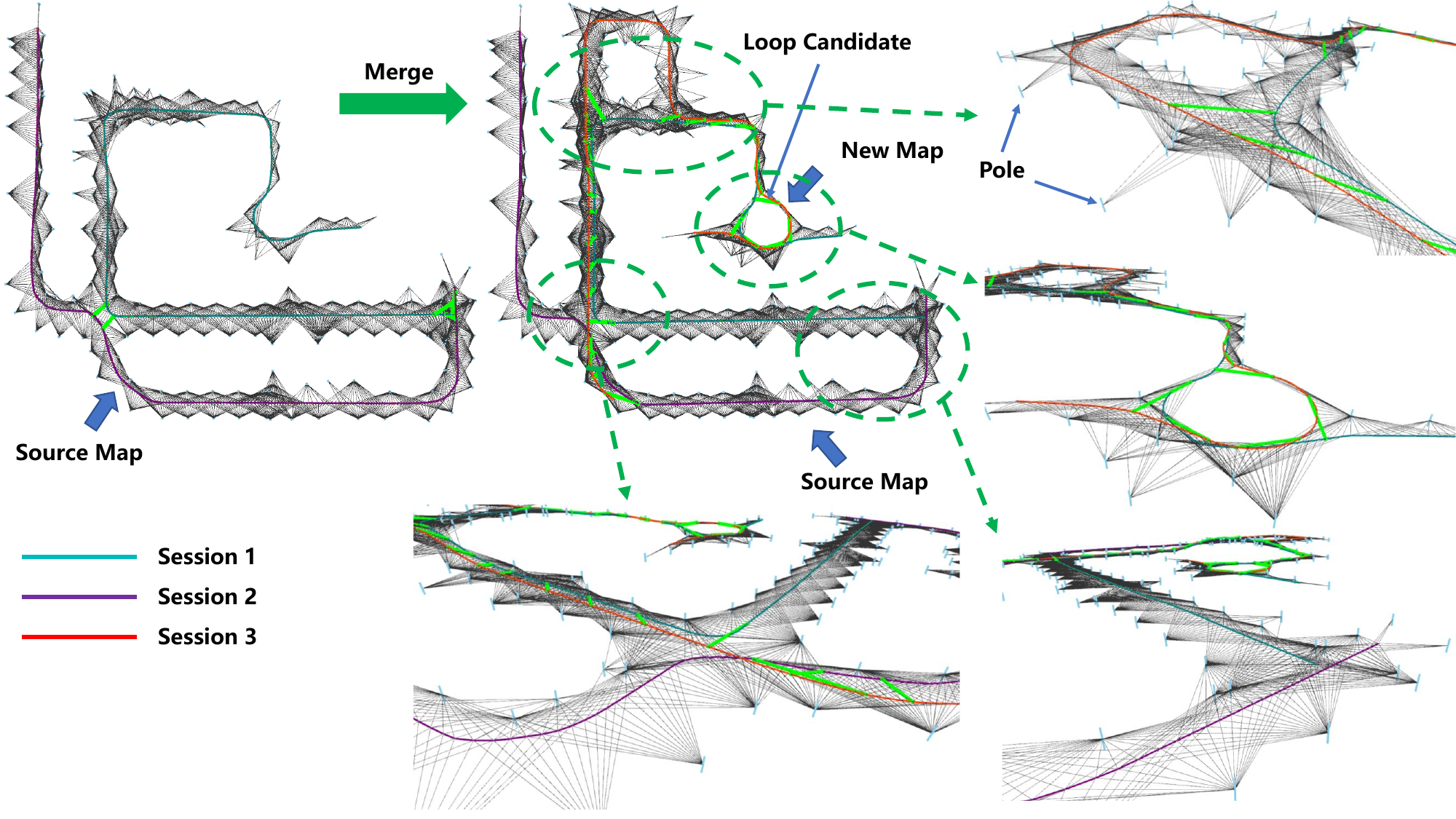}
	\caption{A case of map merging and the co-visible connection with bird's-eye-view on CARLA simulator.  The source map is from two sub-maps (Session 1 and Session 2, top-left) and a new map (Session 3, top-middle) is merged into the source map. The loop candidates (bold green lines) are provided by GraffMatch, and pruned by PCM and all the remaining loop candidates are reliable. The observations between the keyframes and landmarks are rendered as black lines (only poles are visualized for a clearer visualization). }
	\label{figure:FactorGraph}
\end{figure}

In this subsection, we demonstrate that our multi-session mapping framework can merge multiple sub-maps from scratch and construct a globally consistent map. To construct a globally consistent map, the multi-session mapping system is desired to extract loop candidates, eliminate all the false loop candidates and optimize all the keyframes and landmarks. In our experiments, our framework could merge almost all the sub-maps correctly\footnote{A supplementary video presents the global map merging process.}. A case study of map merging is presented in Figure~\ref{figure:FactorGraph}.

\begin{figure}[t]
	\centering
	\includegraphics[width=8cm]{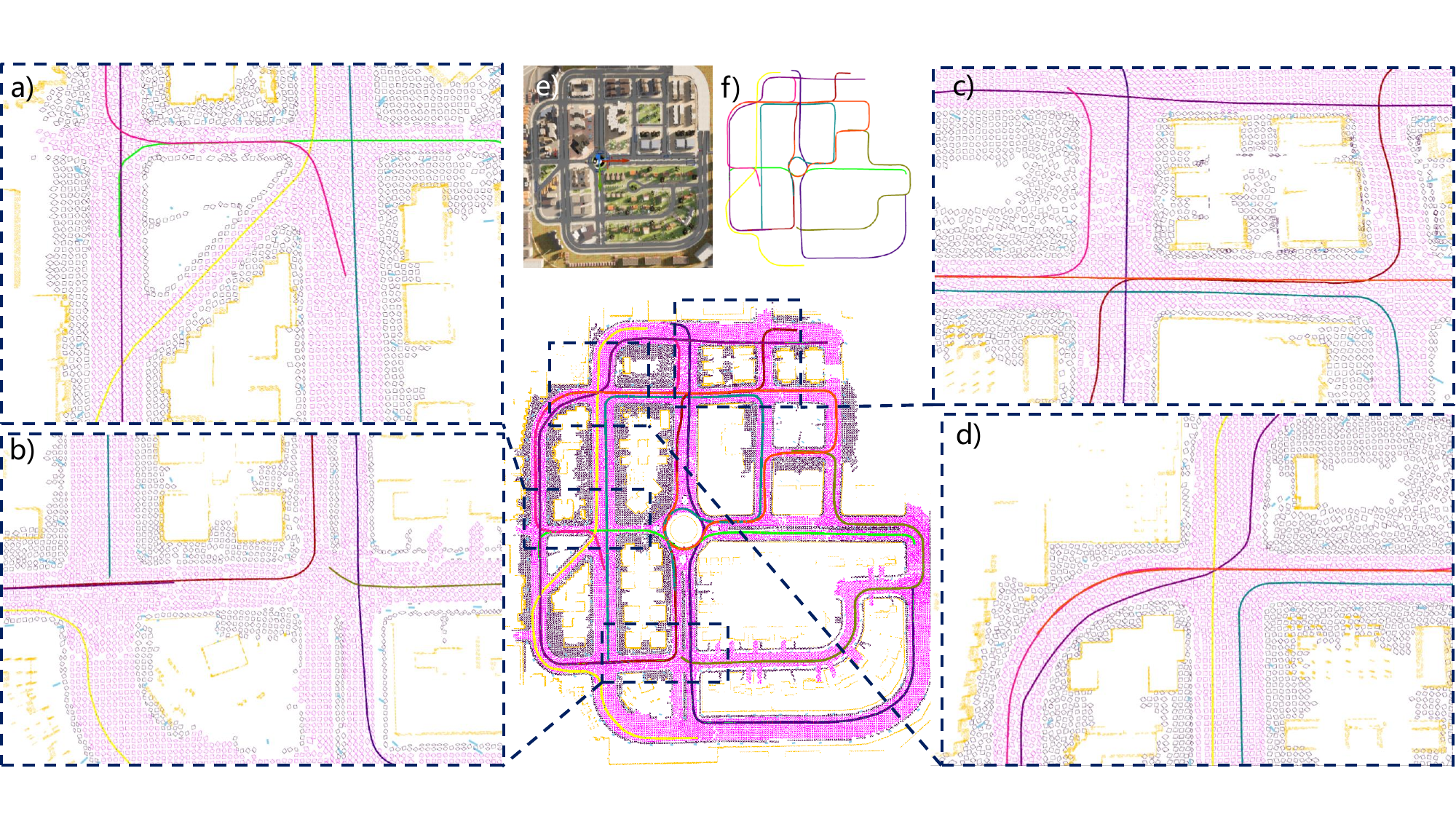}
	\caption{Mapping results with bird's-eye-view on CARLA simulator. The multi-session data is collected in a town using a LiDAR-equipped vehicle. (a)-(d): zoomed views of the semantic lines and planes, as well as colored multi-session trajectories. (e): aerial view of the town seen from the sky. (f): trajectories at different sessions indicated by different colors.}
	\label{figure:CarlaTown}
\end{figure}

To evaluate the mapping accuracy, we use the evaluation results in a LiDAR bundle adjustment (LiDAR-BA) paper~\cite{liu2023large} and the quantitative results of global trajectories are shown in Table \ref{tab:MultiSessionAPE}. Considering that A-LOAM has no loop closure module, we evaluate the global accuracy of SC-A-LOAM\footnote{https://github.com/gisbi-kim/SC-A-LOAM} that has a loop detector \cite{kim2018scan} and a global pose graph optimization as back end. In KITTI Seq.00, our framework provides a more accurate global translation and rotation estimation with our bundle adjustment. In Seq.05, our results are also comparable to other frameworks. However, for Seq.08, we believe that two reasons may have contributed to the inferior results. First, there is not enough overlapping area in Seq.08, resulting in a trajectory dependent solely on odometry. Second, A-LOAM is not as advanced as other LiDAR SLAM systems, and thus estimates a worse trajectory, introducing large errors in our proposed mapping framework. We also present the mapping results on CARLA datasets, shown in Figure~\ref{figure:CarlaTown}.


\subsection{Map Lightweightness}

Despite the consistency of maps, we also evaluate the lightweightness of our proposed map representation compared to the traditional point cloud map. To this end, we conduct experiments on the KITTI dataset and compare the storage requirements of our lightweight map to those of dense point cloud maps that have varying downsampling resolutions $r$. If our map is designed solely for localization without frames or co-visible information, it will include only line and plane landmarks, and we emphasize this with (L) label. The results are summarized in Table \ref{tab:MapStorage}. Our lightweight map requires significantly less storage than the point cloud maps while maintaining map consistency. This feature allows for reduced data transmission loading between the robot and the server, leading to more efficient communication.

\begin{table}[t]
    \centering
    \caption{Lightweight LiDAR Localization on KITTI Dataset: APE ($\degree$/$m$) }
    \resizebox{0.40\textwidth}{!}{
    \begin{tabular}{cccc}
    \toprule
    KITTI Seq. 00 & KITTI Seq. 05 & KITTI Seq. 08 \\
    \midrule
    0.243/0.035 & 0.237/0.045 & 0.376/0.060 \\
    \bottomrule
    \end{tabular}
    }
    \label{tab:LocAccuracy}
\end{table}

\begin{figure}[!t]
	\centering
	\includegraphics[width=\linewidth]{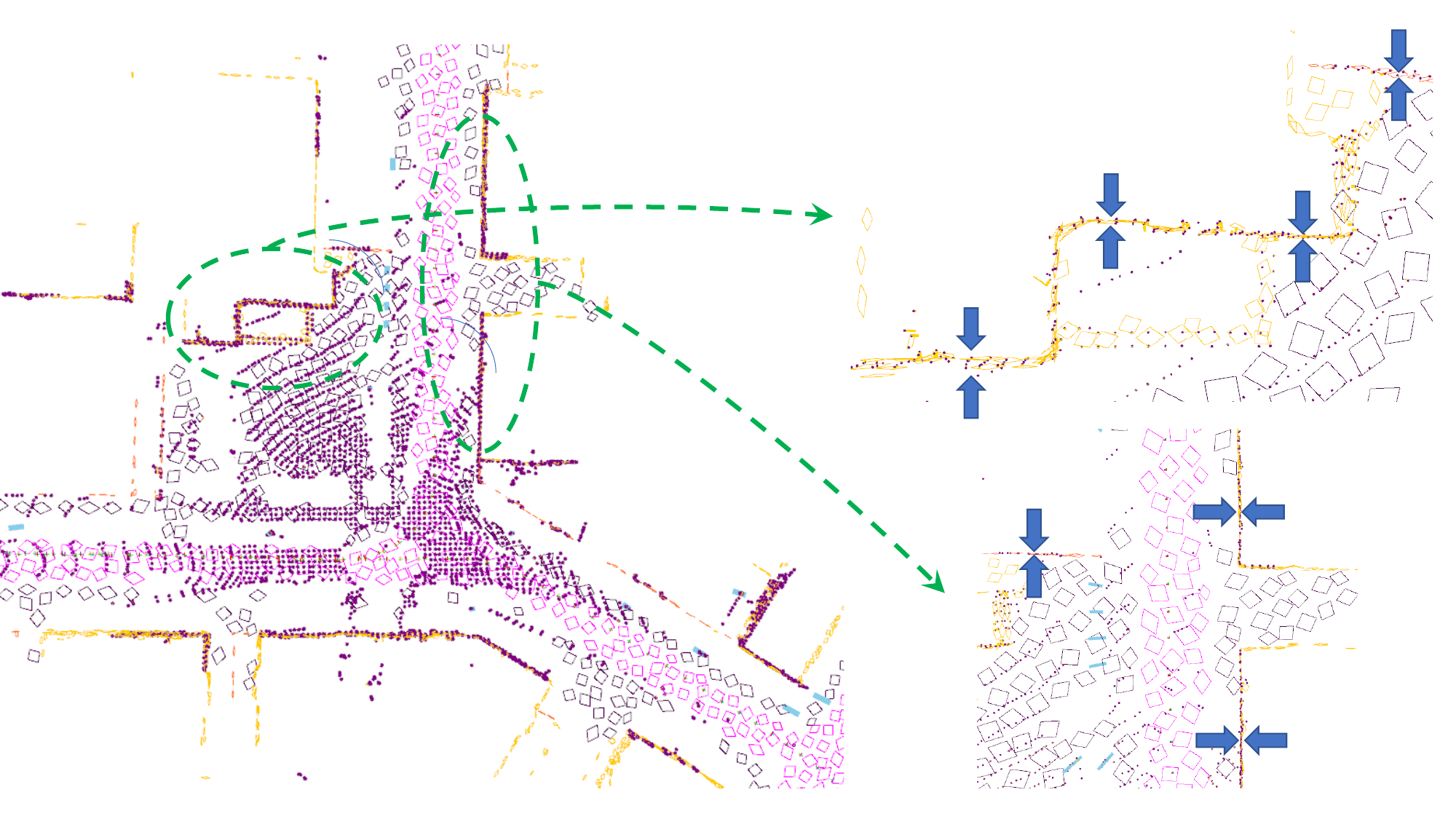}
	\caption{Visualization of online localization on KITTI dataset. Semantics are in different colors in the zoom-in view. The points are registered accurately to the map, demonstrating that our lightweight map is efficient and reliable for online localization. }
	\label{figure:KittiLoc}
\end{figure}

\subsection{Map re-use for online localization}

After obtaining the globally and locally consistent map, we test our lightweight map for online vehicle localization. To avoid the influence of global mapping error, we replace the estimated frame poses with ground truth poses and re-construct the map based on the co-visible information. With this consistent map constructed by the ground truth poses, the global errors for evaluation are only caused by the localization procedure. 

We adopt a semantic-aided point-to-landmarks method to achieve online localization as described in Equation \eqref{eq:OnlineLocalization}.

\begin{equation}
\small
    \begin{aligned}
        \mathop{\mathrm{min}}\limits_{\mathbf{T}^{W}_{L}}
        & \sum_{(k, l)\in \mathcal{M}^{\mathcal{L}}} \rho( \lVert (\mathbf{I} - \mathbf{n}_k\mathbf{n}^T_k)(\mathbf{T}^{W}_{L} \mathbf{p}_l - \mathbf{p}_k) \rVert_{\Sigma_{\mathcal{L}}}^{2}) \\
        & \sum_{(k, l)\in \mathcal{M}^{\mathcal{S}}} \rho( \lVert \mathbf{n}^T_k(\mathbf{T}^{W}_{L} \mathbf{p}_l - \mathbf{p}_k) \rVert_{\Sigma_{\mathcal{S}}}^{2})
    \end{aligned}
    \label{eq:OnlineLocalization}
\end{equation}
where $\mathcal{M}^{\mathcal{L}}$ and $\mathcal{M}^{\mathcal{L}}$ refer to the sets of line landmarks and plane landmarks, respectively, in the map. The variable $\mathbf{p}_l$ denotes the point in the current laser scan, whereas $\mathbf{p}_k$ represents the centroid of the nearest landmark with respect to the current estimated pose $\mathbf{T}^W_L$. All quantitative results for the KITTI dataset are presented in Table \ref{tab:LocAccuracy}. The final localization error is at the centimeter level, which is sufficiently precise for localization-oriented applications, such as autonomous driving in urban environments. Moreover, the average latency for solving the scan-to-map problem \ref{eq:OnlineLocalization} using a single thread on KITTI sequences is approximately 61ms (CPU: Intel i7-8700K), which is sufficient for real-time localization applications. We present a visualization result in Figure~\ref{figure:KittiLoc} to help understand our proposed online localization on the lightweight map.

\section{Conclusion}

In this paper, we propose and validate a multi-session, localization-oriented, and lightweight LiDAR mapping framework in urban environments. The framework includes both global map merging and local refinement, and only uses semantic lines and planes in the pipeline. The generated maps are lightweight compared to point cloud maps and can support online robot localization. There are several promising directions for future work that could improve and extend the proposed framework. Our ultimate goal is to achieve efficient crowd-sourced mapping in city-scale environments.


\bibliographystyle{IEEEtran}
\bibliography{root}

\end{document}